\documentclass[sigconf,screen,authorversion]{acmart}

\usepackage[textsize=tiny]{todonotes}
\usepackage{adjustbox} 
\usepackage{enumitem} 
\usepackage[T1]{fontenc}
\usepackage[ruled,vlined,linesnumbered]{algorithm2e}
\usepackage{booktabs}

\graphicspath{{./}{./Figures}}
\definecolor{mydarkgreen}{RGB}{0, 85, 0}
\definecolor[named]{MyGreen}{HTML}{4bac00}

\newcommand{\hide}[1]{}

\newcommand{\inria}{\affiliation{%
  \institution{Univ. Grenoble Alpes, Inria, CNRS, Grenoble INP, LIG}
  \city{Grenoble}
  \country{France}
}}

\newcommand{\edf}{\affiliation{%
 \institution{Industrial AI Laboratory SINCLAIR, EDF Lab Paris-Saclay}
 \city{Palaiseau}
 \country{France}
}}

\AtBeginDocument{%
  \providecommand\BibTeX{{%
    \normalfont B\kern-0.5em{\scshape i\kern-0.25em b}\kern-0.8em\TeX}}}

\copyrightyear{2023}
\acmYear{2023}
\setcopyright{acmlicensed}\acmConference[SC '23]{The International Conference for High Performance Computing, Networking, Storage and Analysis}{November 12--17, 2023}{Denver, CO, USA}
\acmBooktitle{The International Conference for High Performance Computing, Networking, Storage and Analysis (SC '23), November 12--17, 2023, Denver, CO, USA}
\acmPrice{15.00}
\acmDOI{10.1145/3581784.3607083}
\acmISBN{979-8-4007-0109-2/23/11}

\acmConference[SC23]{The International Conference for High Performance
Computing, Networking, Storage, and Analysis}{November 12--17,
  2023}{Denver, CO}
\begin{document}

\title{High Throughput Training of Deep Surrogates from Large Ensemble Runs}

\author{Lucas Meyer}
\orcid{0000-0001-5386-5997}
\inria
\edf
\author{Marc Schouler}
\orcid{0000-0002-3708-4135}
\inria 
\author{Robert Alexander Caulk}
\orcid{0000-0001-5618-8629}
\inria 
\author{Alejandro Ribes}
\orcid{0000-0001-6141-2146}
\edf
\author{Bruno Raffin}
\orcid{0000-0002-7980-4946}
\inria

\renewcommand{\shortauthors}{Meyer et al.}

\begin{abstract}

  Recent years have seen a surge in deep learning approaches to accelerate numerical solvers, which provide faithful but computationally intensive simulations of the physical world. These deep surrogates are generally trained in a supervised manner from limited amounts of data slowly generated by the same solver they intend to accelerate. We propose an open-source framework that enables the online training of these models from a large ensemble run of simulations. It leverages multiple levels of parallelism to generate rich datasets. The framework avoids I/O bottlenecks and storage issues by directly streaming the generated data. A training reservoir mitigates the inherent bias of streaming while maximizing GPU throughput. Experiment on training a fully connected network as a surrogate for the heat equation shows the proposed approach enables training on 8TB of data in 2 hours with an accuracy improved by 47\% and a batch throughput multiplied by 13 compared to a traditional offline procedure.

\end{abstract}





\maketitle

\section{Introduction}
\label{sec:introduction}

The interest in integrating deep neural networks with traditional numerical simulations has risen in recent years \cite{stevens2020ai,brunton2020machine,karniadakis2021physics,hennigh2021nvidia}. The goal is to accelerate numerical simulations crucial to many scientific and engineering applications \cite{abraham2015gromacs,moin1998direct}. These simulations generally provide a faithful representation of complex physical phenomena at the cost of intensive computation. The numerical simulation, typically a partial differential equation solver, is a process $f$ that takes as input $X$, a vector that encompasses the parameters of the equation, and produces $u_X^t$ a discretized time series of the different time steps of the solution (\autoref{eq:process}):

\begin{equation}
    \begin{aligned}
        f \colon & \mathbb{R}^{d_{\text{in}}} \to \mathbb{R}^{d_\text{out} + 1} \\
        &  X  \mapsto  \{u^t_X\}_{0\leq t < \tau}
    \end{aligned}
    \label{eq:process}
\end{equation}

One of the common machine learning approaches is to design and train a model $f_\theta$, referred to as a \emph{deep surrogate}, that approximates the process $f$. $f_\theta$ is not expected to have the same generalization capabilities as the original solver, but rather, for a given range of parameters, produce much faster simulations.

The deep surrogate can then be used to identify an optimal configuration (e.g. identifying designs in fluid interaction \cite{allen2022inverse}). It can also serve to estimate a probability density function or associated statistics \cite{lee2021scalable}, perform bayesian inference tasks \cite{cranmer2020frontier}, and inverse problems \cite{raissi2019physics}. Machine learning approaches have been reported to be several orders of magnitude faster for such tasks than traditional solvers \cite{wang2019MolDeep,kim19deepfluid,kochkov2021machine}. Deep surrogates are also lighter than traditional solvers and the massive simulation data they produce. These architectures can be seen as efficient compression methods for \emph{post hoc} data analysis and visualization \cite{maulik2020probabilistic,fukami2021machine}. Besides, being developed with deep learning frameworks that support automatic differentiation, these surrogates directly provide different gradients, including the adjoint, valuable for many applications \cite{rackauckas2020universal}.

Generally, the training of these deep surrogates is supervised. It requires the generation of a dataset of simulations. The good generalization capabilities of the surrogate model depend on both the neural architecture and the training dataset \cite{brunton2020machine,satorras2021n,krishnapriyan2021characterizing}. To create a dataset representative of all the richness the process $f$ exhibits when $X$ varies requires executing the solver many times (\emph{i.e.} ensemble runs). The training dataset can quickly grow in size to prohibitive extents,
making it difficult to store on disks and incurring substantial I/O operations, which hinders the training speed.
Additionally, training cannot adapt dynamically the data generation process, the latter being performed \emph{a priori}.

This paper focuses on the combined data generation and training of deep surrogate models. Classical training procedures rely on a fixed dataset that is stored on disk and read to extract batches. However,  because deep surrogates are trained with synthetic data produced by simulation code, the training can be performed online, simultaneously with the data generation. The potential benefits are:
\begin{itemize}
\item {\bf Storage avoiding}. The data are never stored on disk, saving storage space. On supercomputers, storage is often limited, while the increasing size of training datasets exacerbates the demand for storage space and i-nodes.
\item{\bf  I/O bypass}. Directly transferring the data from the simulation to the neural network bypasses storage and circumvents the I/O bottleneck. I/O slows down both the data generation when writing data to disk, as well as training when reading back data from disk. This is a well-known issue in HPC that led to the in situ data processing paradigm \cite{bauer2016situ}. 
\item {\bf Training diversity}. Because the data are produced by a simulation code, the deep surrogate can potentially be trained from an unlimited dataset, exposing the surrogate to more diverse data than with a fixed dataset repeated over several training epochs.  At an equivalent batch count, this higher data diversity can enable online training to converge faster and reach better generalization capabilities compared to offline epoch-based training.
\end{itemize}

The contribution of the present paper is an online and large scale training framework for deep surrogate models called Melissa~\cite{Schouler-JOSS23}. It combines:
\begin{enumerate}
\item  {A multi-level parallelism (parallel solver execution, distributed data parallel training, ensemble run execution), in transit data processing bypassing storage, fault-tolerance for resilience, heterogeneous architecture support, and elasticity for adaptive executions, all features required to ensure efficient training at scale.} 
\item{A training reservoir. A buffer that mitigates the inherent bias in the streamed data caused by the solver internal logic and the availability of computational resources. The reservoir optimizes the throughput of data presented to the GPUs for training while maximizing their diversity.}
\end{enumerate}

We show through experiments that the resulting framework is capable of leveraging multiple levels of parallelism to efficiently train deep surrogates in an online context. It generates and trains a neural network on 8TB of simulation data in less than 2 hours using 5,000 cores and 4 GPUs.
Such a dataset, which could hardly be handled on a cluster in a traditional offline training setting, would require more than 24 hours to be processed by the same number of GPUs. Compared to offline training with multiple epochs on a subset of 100GB,  online training increases by  47\%  the generalization capability of the trained surrogate at an equivalent number of batches.

\autoref{sec:relatedwork} discusses the related work. \autoref{sec:framework} presents the architecture and the developed strategies.  \autoref{sec:experiments} follows with the experiments, while \autoref{sec:conclusion} concludes this paper.


\section{Related Work}
\label{sec:relatedwork}

\subsection{Deep surrogates for numerical simulations}
Recent research on deep surrogate modeling focuses on identifying and training suitable neural architectures for accelerating numerical solutions, while aiming to maintain consistency with physical laws. Most of the deep surrogates found in the literature are trained in a supervised manner from simulation data. The seminal work of Raissi et al. on Physics-Informed Neural Networks (PINNs) could stand as an exception \cite{raissi2019physics}, as they can be trained unsupervised. But  PINNs also benefit from training with simulation data \cite{krishnapriyan2021characterizing,lucor2022simple}. 

When the simulation domain is a regular mesh, time steps $u_X^t$ can be seen as images and convolutional networks can be employed successfully \cite{zhu2019PI-NNSurrogate,kim19deepfluid, ronneberger2015u, wang2020towards, kasim2021building}. The case of irregular meshes can be addressed with specific architectures like Graph Neural Networks (GNNs)\cite{pfaff2020learning,brandstetter2021message} or approached with Fourier neural operators \cite{li2020fourier}. Some have integrated the time dimension using recurrent architectures \cite{tang2020deep} or attention mechanisms \cite{li2023transformer}.

Not only does the space discretization influence the design of the architectures, but the handling of the time dimension also distinguishes \emph{autoregressive} from \emph{direct} models. Autoregressive models mimic the iterative process of traditional solvers, where the current state is used as input to predict the next one (\(f_\theta(u_X^{t-1}) \approx u_X^t\)). One challenge for autoregressive models is the error accumulation along a trajectory, leading to various mitigation strategies \cite{brandstetter2021message,pfaff2020learning,takamoto2022pdebench}. Despite this challenge, the autoregressive approach produces versatile deep surrogates that are, in theory, not limited to the time range they have been trained on.  Direct models produce the state corresponding to the time step provided as input (\(f_\theta(X,t) \approx u_X^t\)). PINNs are an example of direct models that are trained by minimizing the residual error of the PDE at random collocation points \cite{raissi2019physics,sirignano2018dgm,wandel2020learning}.

This paper presents experiments for supervised training of direct deep surrogates. Nonetheless, the presented framework supports the training of any deep surrogate provided it relies on simulation data.  For instance, it has been employed to train autoregressive models at a small scale with a less advanced buffering algorithm \cite{meyer-ICML23}. 

\subsection{Online deep learning}

In the machine learning context, \emph{online training} often refers to long training for which the probability distribution of the presented data shifts over time \cite{parisi2019continual,hoi2021online,sahoo-OnlineDeepLearning-2018}. How this distribution shift occurs depends on the application. Typically, \emph{lifelong training} considers training for very long periods of time as it occurs in recommendation systems. \emph{Streaming learning} characterizes training for which samples arrive continuously and are processed individually. In this paper, the online characterization denotes the simultaneity of the generation of synthetic data, which can be controlled, with the training of the model. 

The probability distribution shift can lead to {\em catastrophic forgetting} where the deep learning architecture trained on recent data sees a deterioration of its capabilities on old data that is no longer or much less present in the training set. 

Deep Reinforcement Learning (Deep RL) is another domain where online training is common. It involves actors that interact with a simulation according to a deep learning architecture that implements the action policy. Several actor instances are executed to produce trajectories that directly feed a learner trained to improve the current policy. {\em Replay buffers} are commonly used as intermediate temporary storage between actors and the learner to mitigate bias and catastrophic forgetting \cite{riedmiller2022collect,fedus-RevisitingFundamentalsExperience-2020,zha-ExperienceReplayOptimization-2019,horgan2018distributed}. 
Actor concurrency also contributes to better data diversity \cite{weng2022envpool}. Solutions developed to orchestrate the online training of the learner on data generated by various actors are specific to Deep RL. They involve specific management of off-policy training, \emph{i.e.} training with trajectories generated under outdated policies. These considerations are irrelevant to the training of deep surrogates because physical laws are constant and never outdated as Deep RL policy can become. This paper reuses the idea of an intermediate buffer but tailored to the context of deep surrogates training. Additionally, it is important to note that even though Deep RL training can require massive distributed resources \cite{silver-MasteringGameGo-2017,nair2015massively,berner2019dota}, the simulation code is not, to our knowledge, computationally intensive according to HPC standards. It runs on a single node, sometimes using a GPU. Thus frameworks for distributed Deep RL, like RLlib \cite{liang2018rllib}, do not face the additional complexity of working with simulation codes parallelized for distributed memory. 

\subsection{Simulation ensemble management}
Efficient management of large ensembles on supercomputers has been a subject of investigation for a long time in applications like sensitivity analysis or data assimilation. The most direct approach relies on files to store intermediate results~\cite{Elwasif-Dakota-CSE2015, Balasu-EnTK-IPDPS2018, Nerger-PDAF-2013,anderson_data_2009,martin-KoraliEfficientScalable-2022}. Thus, each {\em member} of the ensemble, \emph{i.e.} instance of the simulation to run, can be executed independently. Data processing is triggered once all members have been executed. Fault tolerance is easily enabled, but relying on files can impact performance. Using on-node storage sometimes available on supercomputers can contribute to reducing the I/O bottleneck~\cite{Yashiro-1024DA-SC2020}. 

The second standard approach consists in assembling all components of the workflow in a single large MPI application~\cite{Nerger-PDAF-2013,anderson_data_2009}. This is particularly used for data assimilation, which works with cycles of propagation and updates. The members propagate the simulation states for some time steps, then these states are gathered and corrected using observations. Once corrected they are redistributed to the members for them to proceed with the next batch of times steps. If intermediate files are avoided, fault tolerance and load balancing become challenging. These important features, especially when targeting the very large scale, are seldom supported with such an approach. 

Intermediate models have been more recently explored. Members process data online relying on dynamic client/server $N\times M$  data communications \cite{brace2022coupling,terraz2017melissa,friedemann-melissaDA:2022}. These intermediate models keep the best of both worlds: the efficiency of a file-avoiding solution while retaining the necessary flexibility to support fault tolerance, load balancing and some elasticity. The framework presented in the paper adopts this approach, extending the Melissa framework initially developed for sensibility analysis~\cite{terraz2017melissa}.


\subsection{Task and workflow}
Ensemble runs are a specific type of workflow, often developed with distributed task-based environments or workflow managers. Examples of these environments and managers include Ray, Dask, Parsl, Pycompss, RADICAL-Cybertool, qgc-pilot ~\cite{rocklin-DaskParallelComputation-2015,babuji-parsl-2019,moritz_ray_2018,tejedor-PyCOMPSsParallelComputational-2017, balasubramanian-RADICALCybertoolsMiddlewareBuilding-2019, bosak-VerificationValidationUncertainty-2021}. Few are actually capable of enabling $N\times M$ dynamic connections between legacy MPI parallel tasks while ensuring fault tolerance. 

The possibilities of exascale computing to open new scientific opportunities through large ensembles has been stressed early for molecular dynamics~\cite{Pronk-copernicus-gmx-SC11}. Deep surrogates are relevant in pure numerical schemes, but also in workflows combined with other scientific instruments~\cite{aad2022atlfast3,alexander-CodesignCenterExascale-2021}. Some are reporting gains of several orders of magnitude\cite{wang2019MolDeep,lee2019deepdrivemd} when assessing globally the cost of deep surrogate training and the gains of using the surrogate versus the original simulation. The basic workflow sequences two steps: 1) surrogate training, 2) surrogate inference for addressing the target problem, potentially combined with some simulation runs when higher precision is needed. But some are pushing the logic one step further fusing these two steps into a single {\it adaptive ensemble run} where a steering logic, relying on shallow or deep learning, tries to improve the global workflow efficiency~\cite{ward2021colmena,zamora-ProximaAcceleratingIntegration-2021,balasubramanian-EnTk-2020}. In this paper we focus on the deep surrogate
training process (step 1), but our approach has all the necessary flexibility to be used in the fused workflow.   

To conclude this section, we mention  emerging approaches combining  ensemble runs and deep learning, like simulation-based inference~\cite{cranmer2020frontier,baydin2019etalumis} or  simulation intelligence~\cite{lavin-SimulationIntelligenceNew-2022},
that further stress the growing potential of deep surrogates and online training.

\section{Framework}
\label{sec:framework}

The presented framework aims to optimize the throughput of data generation, transmission, batch creation, and distributed training for deep learning at a large scale (thousands of CPUs and multiple GPUs). At its core, the design leverages the stochastic nature of gradient descent, where the data presented to a Neural Network (NN) for training are sampled according to a given density distribution (usually uniform) and not strictly ordered. The framework exploits these loose synchronization and data ordering requirements to improve large-scale resource usage. In the following, \autoref{sec:arch} presents the framework's components and their assemblies, then \autoref{sec:data} provides the details of data management and buffering algorithms.

\subsection{Architecture}\label{sec:arch}

\begin{figure}[h]
    \includegraphics[width=\columnwidth]{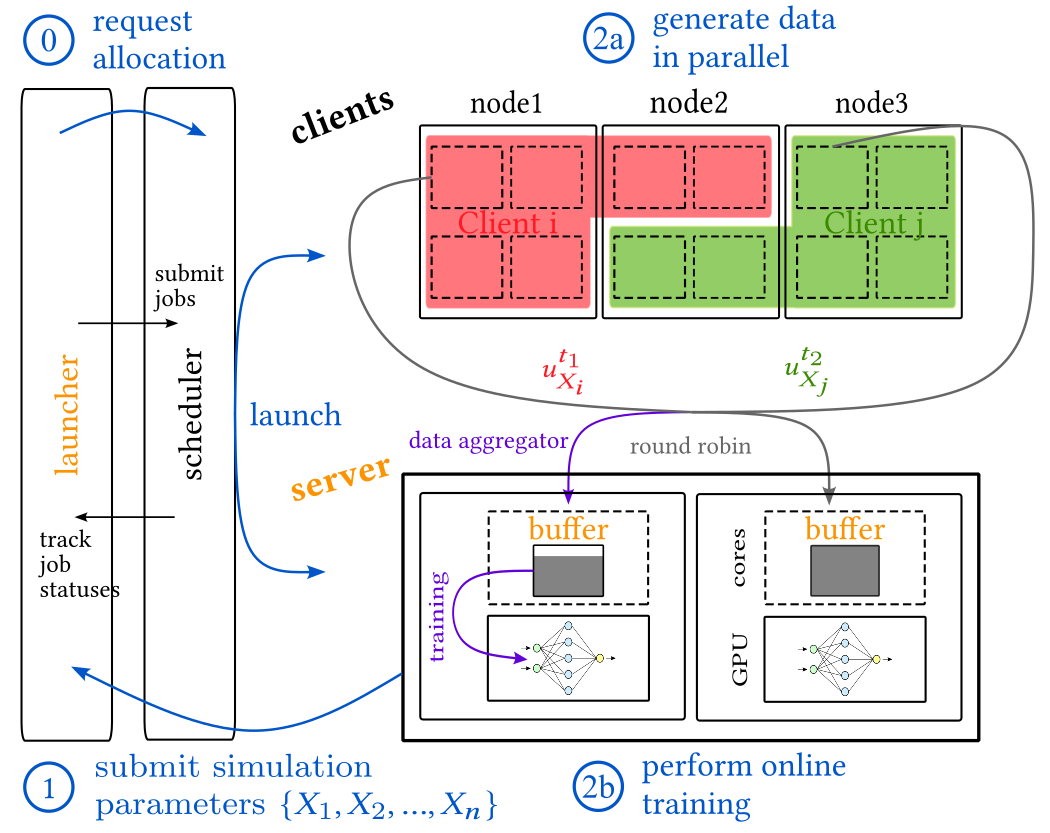}
    \caption{
        Framework architecture. Core components are highlighted in \textcolor{ACMOrange}{orange}. The different steps of the workflow are represented in \textcolor{ACMDarkBlue}{blue}. The data generation \textcolor{ACMDarkBlue}{(2a)} and the training \textcolor{ACMDarkBlue}{(2b)} occur simultaneously. Here, 2 clients, \emph{i.e.} 2 simulation instances of respective parameters \textcolor{ACMRed}{\(X_{i}\)} and \textcolor{MyGreen}{\(X_{j}\)}, run on 6 cores each, spanning over a total of 3 nodes. Training is performed by the server with distributed data parallelism on 2 GPUs. As soon as time steps (e.g. \textcolor{ACMRed}{\(u_{X_{i}}^{t_1}\)} and \textcolor{MyGreen}{\(u_{X_{j}}^{t_2}\)}) are computed by the clients, they are streamed to the server. On each process of the server a \textcolor{ACMPurple}{data aggregator thread} polls for new data to store in the buffer. Concurrently, the \textcolor{ACMPurple}{training thread} extracts batches from the buffer and proceeds with training.
    }
    \label{fig:melissa_architecture}
\end{figure}

The framework architecture relies on a client/server model extended to the parallel case where both the client and server are programs with potentially different levels of parallelism (\autoref{fig:melissa_architecture}). No intermediate file is required as all data exchanges take place through direct memory-to-memory communications between the clients and the server.

This client/server architecture improves the application modularity. Because the connection between a client and the server is dynamic, a client can be stopped (voluntarily or not) and started anytime. A client failure does not lead the server to failure, providing a sound base to support an efficient fault tolerance protocol. The number of running clients can evolve with time according to the resources available on the supercomputer, making the application elastic.

Each \textbf{client} runs an instance of the simulation code $f$ with different input parameters $X$. The simulation is often an MPI+X parallel code running on several cores and nodes. As soon as a client produces a new time step $u_X^t$, this one is sent to the server via the API. Because clients run as independent executables, each one can use different types or amounts of resources.

The \textbf{server} is in charge of training. It is an MPI code relying on distributed data parallelism for parallel training. All MPI processes run an identical copy of the NN, but each one trains it with different data. After each batch backpropagation, the locally computed vector of weight updates is all-reduced between all processes and applied to each local NN copy to keep them identical~\cite{li2020pytorch}. This is today the standard parallelization approach for training, capable of using thousands of GPUs if learning rate and batch sizes are managed properly~\cite{goyal2017accurate,mikami-2018}. The framework has not yet been experimented with model parallelism \cite{jia2019beyond,Xu_2020_CVPR}. Model parallelism is often combined with data parallelism when the NN cannot fit into the memory of one GPU. This would require splitting the NN and the batches to enable pipeline parallelism for instance as well as making parallel all-reduces per group of GPUs managing the same sub-part of the NN \cite{shoeybi-MegatronLMTrainingMultiBillion-2020}. In the context of our framework, changes are expected to be mainly limited to the code of the training thread. Today, DL frameworks like Pytorch or TensorFlow, on which we rely, provide extensions to ease the deployment of multi-GPU model parallelism.

Each server process runs two threads. The {\bf data aggregator thread} manages connections to clients, receives data and stores these data into the training buffer. The second thread, the {\bf training thread}, reads data from the training buffer to build a batch, feeds the GPU with it and performs the forward and backward passes through the NN. An all-reduce operation amongst the different training threads aggregates the gradients to update the network weights. Finally, each thread copy of the network is updated locally before repeating the process with a new batch. 

The data aggregator thread controls the \textbf{experimental design}. Methods currently supported to draw the parameters $X$ for each client include the traditional Monte Carlo method, Latin hypercube and Halton sequence. Because the server drives the training progress, the experimental design could be made adaptive to support \emph{active learning} strategies.

There is one \textbf{training buffer} per server process. This is a thread-safe data container of fixed capacity shared between the aggregator and the learner thread. This buffer, discussed in detail in \autoref{sec:reservoir}, is a critical component to balance the quality and speed of training.

The \textbf{launcher} orchestrates and monitors the workflow. The launcher interacts with the supercomputer batch scheduler to start clients or server jobs, monitor their progress, kill some of them or restart them in case of failure. The launcher first starts the server job. Next, the server forwards to the launcher requests for executing client instances.  The launcher takes care of building the associated jobs and submits them to the batch scheduler. The launcher's default behavior is to request one resource allocation for the server and one per client. But this approach shows limitations in two cases. First, when the client run is very short or just requires a few cores, the scheduling overheads dominate leading to idleness on the server side.  Next, as all jobs are independent their start time depends on resource availability and may also lead to server idleness when the number of running clients is too low. To mitigate these issues, the framework provides a schedule-in-schedule approach where a larger resource allocation is requested (or several large ones) and jobs are scheduled into this allocation. Currently, the framework supports the Slurm \cite{yoo2003slurm} and OAR \cite{capit2005batch} schedulers. Integration of workflow schedulers like RadicalPilot \cite{merzky2015radical} or Flux \cite{ahn2020flux}, to directly and efficiently take care of this two-level scheduling scheme, is planned.

The workflow is heterogeneous, usually running clients on CPU nodes and the server on GPU nodes, each type of node being managed independently through two different scheduling queues. As highlighted by experiments (\autoref{sec:experiments}), much fewer GPUs are usually needed compared to the number of CPUs. As the server starts first, it is natural to request first a reservation of GPU nodes and next CPU nodes for clients.  However, the CPU partition was significantly more loaded than the GPU one, leading to a server staying idle for long periods while waiting for CPU resources. We thus had to reverse the reservation scheme, requesting first CPU resources, and, once available, GPU resources for the server. This proved to be the most economical approach to preserve our compute hour budget. Notice that schedulers like Slurm can support directly such heterogeneous jobs, but we were asked not to use this feature as it apparently could affect the resource allocation efficiency.

The framework is \textbf{fault-tolerant}. The server watches for unresponsive clients and asks the launcher to properly kill and restart faulty ones. The server maintains a log of received messages per client, so in case of client restart, already received messages are discarded.  If the client simulation code supports checkpointing, it can be enabled so the client will restart from the last checkpoint only. The server is regularly checkpointed. If a server failure is detected by the launcher, it first kills all running clients and next restarts a new server instance from the last checkpoint. This server will request the launcher to restart the necessary client instances. When the launcher fails, the currently running clients continue until completion, after which the server checkpoints and stops. It has then to be restarted manually.

The framework, Melissa,  is open source\footnote{\url{https://gitlab.inria.fr/melissa/melissa}}. All the stochastic components (\emph{i.e.} the network weight initialization, the simulation parameter sampler, and the training buffer) are seeded for reproducibility purposes.

A minimalist API for C, Fortran, and Python enables to instrument the simulation code for the clients. A first call is required to connect the client to the server (\emph{init\_communication}). A \emph{send} is issued to transfer time steps $u_X^t$ as soon as computed. Eventually, a client calls \emph{finialize\_communication} to signal the server that no more data will be sent before disconnecting. We also provide a PDI plugin to interface with PDI instrumented simulation codes \cite{roussel-PDI-2017}.

The launcher and server are developed in Python. Transport layer relies on ZMQ \cite{hintjens2013zeromq}. We are considering adding ADIOS2 \cite{godoy2020adios} for gaining on data handling flexibility and better use high performance networks.

Regarding training, the framework supports the PyTorch and Tensorflow libraries. The training thread embeds a classical training loop where the main difference is the data source that relies on the training buffer instead of files. To ease the user transition from offline training for prototyping to online training, the buffer has been abstracted through the classical Tensorflow and Pytorch Dataset classes.

\subsection{Data management}\label{sec:data}

\subsubsection{Data diversity}

In classical offline training, the gradient descent expects batches built by uniformly sampling the fixed dataset, which can easily be done as the full dataset is available upfront \cite{bottou2018optimization}. The training process of our framework being online, it leads to inherent sources of data bias. Bias in the data is known to be detrimental to the quality of training. For instance, catastrophic forgetting characterizes network performance decrease on previously seen data when the training data distribution changes~\cite{kirkpatrick2017overcoming}. The sources of {\bf workflow bias} caused by online training are of three different categories:

\begin{itemize}
    \item{\emph{Inter-simulation}: The computational resources are finite. At any time, only \emph{c} different clients can run simultaneously. The data produced by the \emph{c} simulations may not present all the diversity the process $f$ exhibits.}
    \item{\emph{Intra-simulation}: Most of the time, the simulation codes executed by each client produce time steps in fixed increasing order. Each time step $u^t_X$ is sent in the same order to the server as soon as computed. Obviously, time steps that have not been computed yet are not available for training.}
    \item{\emph{Memory budget}: The training  buffer  has a fixed capacity $n_c$. At any time of the experiment, it can store only a subset of all the data received so far, $n_c \ll |\mathcal{X} \times \mathcal{T}|$  where $|\mathcal{X} \times \mathcal{T}|$ represents the cardinality of the set of data generated by the clients during the experiment.}
\end{itemize}

The following details the different techniques the framework implements to ward off these biases and enable high-quality training and efficient use of resources.

\subsubsection{Data distribution}

The different GPUs involved in the training require the field $u_X^t$ associated with each time step they receive to be complete. Each parallel client produces a time step partitioned across its different processes. The assembly of these parts is performed in situ on each client through an MPI gather on rank zero. Because training can usually be performed with data at a lower precision than the one produced by the solver, they are gathered and then converted, typically from 64 to 32 bits. Thus we avoid overloading the server with all these preprocessing tasks that need to be performed for each time step produced by each client.

To reduce inter and intra simulation bias, each client connects to all the ranks of the server and distributes the produced time steps $u^t_X$ across all GPUs in a Round-Robin fashion. The destination of the first time step is chosen according to the client id to limit having all clients sending the same time step to the same GPU. This enforces the balance of data for data parallel training.

\subsubsection{Training buffer}
\label{sec:reservoir}



The training buffer is a fixed-size data container that has the dual role of accumulating a certain number of time steps to ensure batches contain well-mixed data to reduce workflow bias, and amortizing (up to a certain limit) discrepancies between data production and consumption to reduce resource idleness. The {\bf Reservoir} algorithm we propose for managing this buffer is a key contribution of this paper.  For comparison purposes, we also present two other strategies, FIFO and FIRO. Prior work only considered FIRO implementation \cite{meyer-ICML23}, which is shown here to fail in optimizing GPU usage.

The classical  {\bf FIFO buffer (First In, First Out)} corresponds to the streaming case, where data are batched for training according to the order they are received. Each data produced is seen once, and only once, during training. Batch extraction is enabled as soon as the buffer can provide one.  Compared to pure streaming, buffering provides some slack to keep the consumer, the learner thread, busy as long as batches are available in the buffer, even if data production is stopped or reduced for a while. Data production is suspended when the FIFO buffer is full. The FIFO buffer is thus simple to implement as a queue but does not enable mixing data beyond what the server receives.

The {\bf FIRO buffer (First In, Random Out)} behaves very similarly to FIFO, with data evicted upon reading, except that these data are extracted from random positions to build less biased batches. Additionally, batches can only be extracted if the buffer is filled beyond a given \emph{threshold}, again to reduce batch bias. The threshold is set to zero once data production is over to enable consuming the last produced data. FIRO is implemented as a list container. Newly received samples are appended at its end.

\begin{algorithm}
    \DontPrintSemicolon
    \SetKw{Int}{int}
    \SetKw{List}{list}
    \SetKw{Bool}{bool}
    \SetKw{Lock}{lock}
    \SetKw{Delete}{delete}
    \SetKw{Wait}{wait}
    \SetKw{In}{in}
    \SetKw{Not}{not}
    \SetKwFunction{Put}{put}
    \SetKwFunction{Get}{get}
    \SetKwProg{Fn}{Function}{:}{}
    \SetKwComment{Comment}{  \# }{}
    \KwData{\Int{capcity}, \Int{threshold}, \List{seen}, \List{not\_seen}, \Bool{is\_reception\_over}, \Lock{lock}}
    \Fn{\Get{}}{
        lock.acquire()\;
        \While{seen.size + not\_seen.size $\leq$ threshold}{
            \Wait{} \Comment{Ensure there are enough data}\;
        }
        index = random.select(seen.size +  not\_seen.size)\;\label{algo:index}
        \If{index \In{not\_seen.indices}}{
            item = not\_seen[index]\;
            \Delete{\ }not\_seen[index]\;
            \If{\Not{is\_reception\_over}}{
                seen.append(item)\;
            }
        }
        \Else{
            item = seen[index]\;
            \If{is\_reception\_over}{
                \Delete{\ }seen[index]\Comment{Empty the buffer}\;
            }
        }
        lock.release()\;
        \Return{item}\;
    }
    \Fn{\Put{item}}{
        lock.acquire()\;
        \While{not\_seen.size $\geq$ capacity}{
            \Wait{} \Comment{Block until one element gets seen}\;
        }
        \If{not\_seen.size + seen.size $\geq$ capacity}{
            index = random.select(seen.size)\;
            \Delete{\ }seen[index] \Comment{Evict one seen element}\;
        }
        not\_seen.append(item)\;
        lock.release()\;
      }
      \caption{Reservoir}
      \label{algo:reservoir}
\end{algorithm}

The  {\bf Reservoir buffer}  (Algorithm~\ref{algo:reservoir}\hide{autoref not working properly withitsalgo}) enables data to be seen more than once to reduce consumer idleness in case of underproduction while giving priority to storing newly produced data over already seen ones. The Reservoir distinguishes the new unseen data from the ones already selected in a previous batch. When receiving new data while the buffer is full, a seen example is randomly evicted to make room for the incoming one. When building a batch, the elements are uniformly selected one by one among the seen and unseen elements in the buffer. Each selected unseen data is then moved with the seen ones. This Reservoir ensures that batches are well mixed;  that data production can proceed as long as the buffer is not full of unseen samples, avoiding discarding any unseen data; and that data consumption is never locked, once the threshold is passed, as new batches can be built from already seen data. The Reservoir also has a \emph{threshold} of minimum stored data before drawing batches. This ensures that the first received time steps are not over-represented in the first batches of the training. It also ensures the buffer has a minimum population to create diverse batches. When all the simulation data have been generated the blocking related to the threshold is lifted and the buffer state is updated. Batches can then be freely drawn from the buffer, regardless of its population size or the seen nature of the samples, until it finally empties out. When the reception is over and the buffer is empty, the training terminates. Notice that the amount of buffer space for unseen data is regulated by the incoming flow of new data, avoiding the static split that a dual buffer would require. Notice also that batch selection is with replacement. A selection without replacement would require a slight modification but would increase the selection cost. Experiments in \autoref{sec:experiments} show the effectiveness of this training Reservoir.

For the Reservoir, data production pushes out data from the buffer. This can potentially lead to catastrophic forgetting as data from the earlier simulations get evicted.  The expected residency time of a sample  $u_X^t$ in the buffer is about $n_c$, the buffer capacity or the number of unseen elements for the training buffer  (\autoref{proof:item_life_expectancy}). Because of the online setting, the framework relies on the experimental design to ensure that the parameter space is well sampled to continuously populate the training buffer with diverse data. Experiments confirm that this avoids catastrophic forgetting. In Deep RL, a secondary replay buffer can be used to limit catastrophic forgetting, using the \emph{Reservoir Sampling}  algorithm \cite{efraimidis2006weighted,zhang-FrameworkDualReplay-2019}.  Reservoir sampling is a randomized algorithm to populate a $k$-size buffer from a stream of data guaranteeing that at any time $\tau$ the buffer is filled with $k$ distinct elements uniformly sampled from the $\tau$ data received. Directly using this algorithm for online training would be counterproductive as it would waste the produced data not selected for inclusion in the buffer. As a secondary buffer, it would compete with the main buffer for the node memory.

\section{Experiments}
\label{sec:experiments}

Experiments\footnote{Experiments are available for reproducibility at \url{https://gitlab.inria.fr/melissa/sc2023}} consist in training deep surrogates of a heat-equation solver (\autoref{sec:use-case}). We first assess the training throughput for the FIFO, FIRO and Reservoir buffers with a single GPU in \autoref{sec:exp:throughput}, before comparing the obtained training quality in \autoref{sec:exp:training_quality}. Multi-GPU training is considered in~\autoref{sec:multigpus}. \autoref{sec:offline} finishes with a comparison between online and offline training at a larger scale.

\subsection{Equation and deep surrogate architectures}
\label{sec:use-case}


The experiments consider the classical heat equation on a 2D rectangular domain (\autoref{eq:heat-pde}):
\begin{equation}
    \left\{\begin{array}{l}
         \frac{\partial T}{\partial t} = \alpha \nabla^2 T, \\
         T(x,\, y,\, 0) =T_{\text{IC}}, \\
         T(0,\, y,\, t) = T_{x_1},\, T(L,\, y,\, t) = T_{x_2}, \\
         T(x,\, 0,\, t) = T_{y_1},\, T(x,\, L,\, t) = T_{y_2},
    \end{array}\right.
    \label{eq:heat-pde}
\end{equation}
where $T(x,y,t)$ denotes the field temperature, $\alpha$ the thermal diffusivity and $(T_{\text{IC}}, T_{x_1}, T_{y_1},  T_{x_2}, T_{y_2})$ the initial and 4 boundary temperatures. The solution is approximated with an in-house solver that implements a finite difference method with an implicit Euler scheme. The temperature field $T(x,y,t)$ is discretized on a $1000\times1000$ Cartesian grid, and generated for $100$ time steps representing $\Delta t=0.01$ second each. The thermal diffusivity is fixed to  $\alpha = 1$ $\text{m}^2.\text{s}^{-1}$. The solver code is written in Fortran90 and parallelized with MPI according to classical 2D domain partitioning.

The surrogate is trained to directly predict the temperature field \(u^t_X=T(x, y, t)\) given the input \((X,t)\), \(X=(T_{\text{IC}}, T_{x_1}, T_{y_1},  T_{x_2}, T_{y_2})\). Training data are generated from solver runs taking as input 5 temperature parameters \(X\) (initial and boundary conditions) randomly sampled in \(\left[100, 500\right]\)K.

The deep surrogate architecture is a multilayer perceptron of  514M parameters (for reference the language model Bert-Large has about  340M parameters \cite{kenton2019bert}), consisting of an input layer of 6 neurons, 2 hidden layers of 256 neurons with ReLU activation and an output of 1M neurons. It is trained using Adam optimizer with a starting learning rate of $1\text{E}^{-3}$.

\subsection{Computational resources}

Experiments are performed on the Jean-Zay supercomputer ranked in the first half of the top500.org as of November 2022. The machine has one GPU and one CPU partition. Jean-Zay's CPU partition consists of 2 Intel Cascade Lake 6,248 processors with 20 cores at 2.5 GHz for a total of 40 cores per node. The  GPU partition provides nodes accelerated with 4-GPU V100 32 GB. These GPU nodes came with 40 CPU and 160 GB of RAM. Jean-Zay's nodes are connected with Intel Omni-Path (100 GB/s). IBM Spectrum Scale (ex-GPFS) parallel file system with SSD disks (GridScaler GS18K SSD) manages the storage.

\subsection{Throughput}
\label{sec:exp:throughput}

The experiment illustrates how the different buffer implementations impact the throughput during training on 1 GPU. It considers a dataset of 250 runs of the heat equation solver, accounting for a total of 25,000 time steps. These data are generated by the framework running concurrently 100 clients on 50 nodes. Each instance of the solver executed by the clients is parallelized over 20 cores. Due to the limited support for heterogeneous jobs (CPU/GPU) on the machine, the framework adopts the submission process described before in \autoref{sec:arch}.  It submits three series of clients. First, it launches 100 simulations. Once the clients have terminated their executions, it launches the second series of 100 and finally the third series of the 50 remaining simulations.   The throughput assesses the capacity of the framework to quickly provide training data to the GPU. The throughput is expressed as the number of samples per second processed by the GPU, computed on the learning thread over 10 successive batches every 10 batches. The batch size is fixed to 10 samples, with one sample being the time step $u_X^t$  of one simulation associated with its 6  input parameters $(X,t)$.

For this experiment only, as we are solely considering throughput, the performance of the trained network is never evaluated on a validation dataset. Performing the evaluation on the training thread would stall the data consumption of training batches, thus reducing throughput. Evaluation can either be excluded from the throughput measurement (as in the next experiments), or performed by using a dedicated GPU not involved in training.

The three training buffers described in \autoref{sec:reservoir},  \emph{FIFO}, \emph{FIRO} and \emph{Reservoir}, are evaluated. FIFO yields samples as they are received from the clients.  FIRO and Reservoir have a fixed capacity of 6,000 samples (roughly a fourth of the whole simulated data) with a  threshold set to 1,000 samples. These parameters remain the same in all the experiments.

\begin{figure}[h]
    \includegraphics[width=\columnwidth]{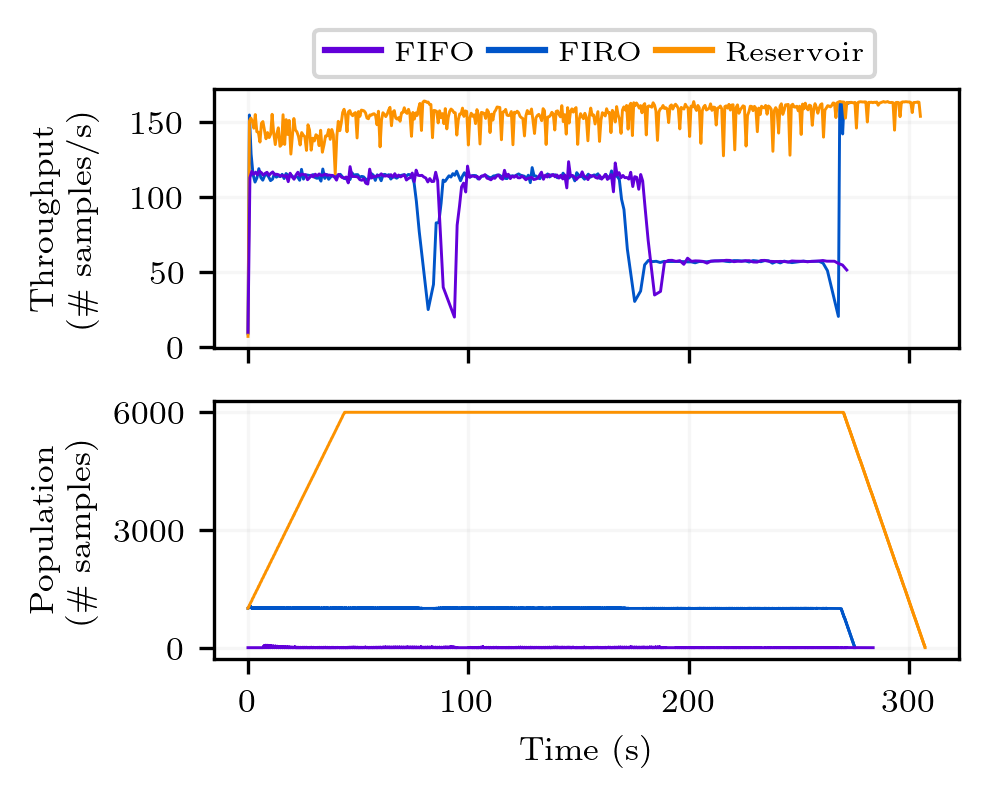}
    \caption{Reservoir population and throughput for different implementations. Each one handles a total of 25,000 time steps. The data are generated by successive series of 100, 100, and 50 clients running concurrently.}
    \label{fig:throughput}
  \end{figure}

\autoref{fig:throughput} shows the evolution of the throughput with respect to time. The Reservoir provides the highest throughput of the three implementations. It appears that both FIFO and FIRO are sensitive to the flow of incoming data. Their throughput drops briefly at 100 and again at 200 seconds. After 200 seconds, their throughput is half what it was before. The times these drops occur coincide with the transitions between series of client submissions performed by the launcher. A bit before 100 seconds, all the running clients have terminated generating the data of the first 100 simulations. There is a delay before the clients start running the next 100 simulations and streaming the data they generate to the server. Similarly, there is a delay between the execution of the second series of 100 simulations and the execution of the last series of 50 ones.  The amount of generated data is also halved during this last series running 50 instead of 100 concurrent simulations. Because FIFO yields data as they arrive these changes in data production result in drops in the throughput. The FIRO shows similar patterns, but it occurs sooner as the data consumption is stopped when reaching the threshold value and not when emptying the buffer as for FIFO.

\autoref{fig:throughput} bottom shows the FIFO and FIRO buffer populations stay around their minimum, respectively 0 and 1,000 set by the threshold value. Because data are consumed faster than produced their throughput is virtually the one of the data generation by the clients. At the end of the experiment the FIRO throughput rockets up. At this time, the data production by the clients has terminated and the blocking threshold has been released. Samples can thus be freely drawn to form batches. The FIRO population, which is necessarily higher than the fixed threshold of 1,000 samples, can quickly yield 100 batches. Because the Reservoir evicts on writing rather than on reading like FIRO,  its population increases quickly to its maximum, while FIRO's population remains around the threshold value, even when data consumption equals or overruns production as in this case. We can expect this enables Reservoir batches to present a higher diversity as they are sampled from a  larger population.

  \begin{figure}[h]
    \includegraphics[width=\columnwidth]{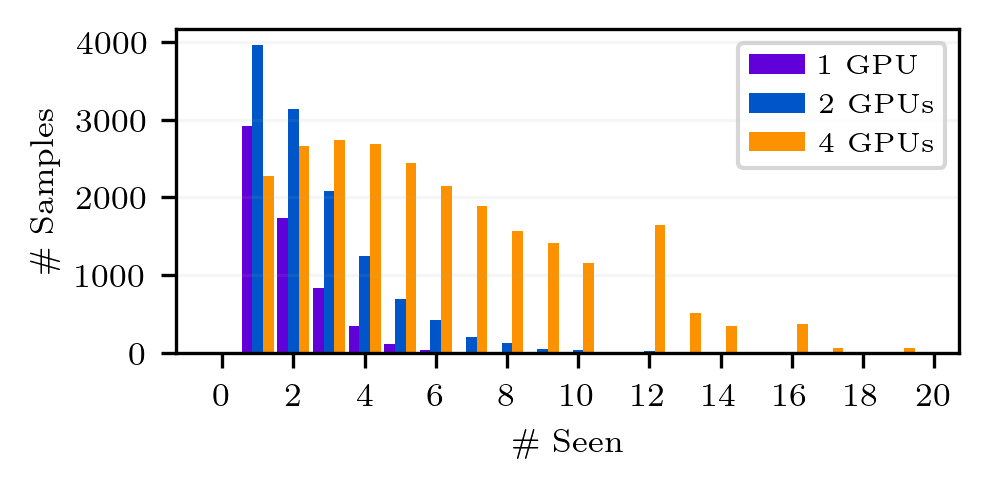}
    \caption{Number of occurrences of simulation time steps in batches for the Reservoir for different numbers of GPUs.}
    \label{fig:seen}
  \end{figure}

The Reservoir throughput remains constantly higher than the ones of FIFO and FIRO. The Reservoir manages to provide a high throughput by repeating samples, which erases the gaps between production and consumption rates. The visible high frequency fluctuations are not yet understood.
\autoref{fig:seen} presents a  histogram of the number of times samples have been repeated for the Reservoir. Most of the samples have been seen in batches a couple of times, and at most, though rarely, 8 times during training. Thus, a few repetitions are enough to increase the throughput by 50\%.

By providing higher throughput, the Reservoir maintains the GPU active, whereas for FIFO and FIRO implementations it can be idle, waiting for data to come. To be useful, this extra activity in training on more batches with already seen samples must result in increased generalization capabilities, which should appear as a lower validation loss.

\subsection{Training quality}
\label{sec:exp:training_quality}

\begin{figure}[h]
    \includegraphics[width=\columnwidth]{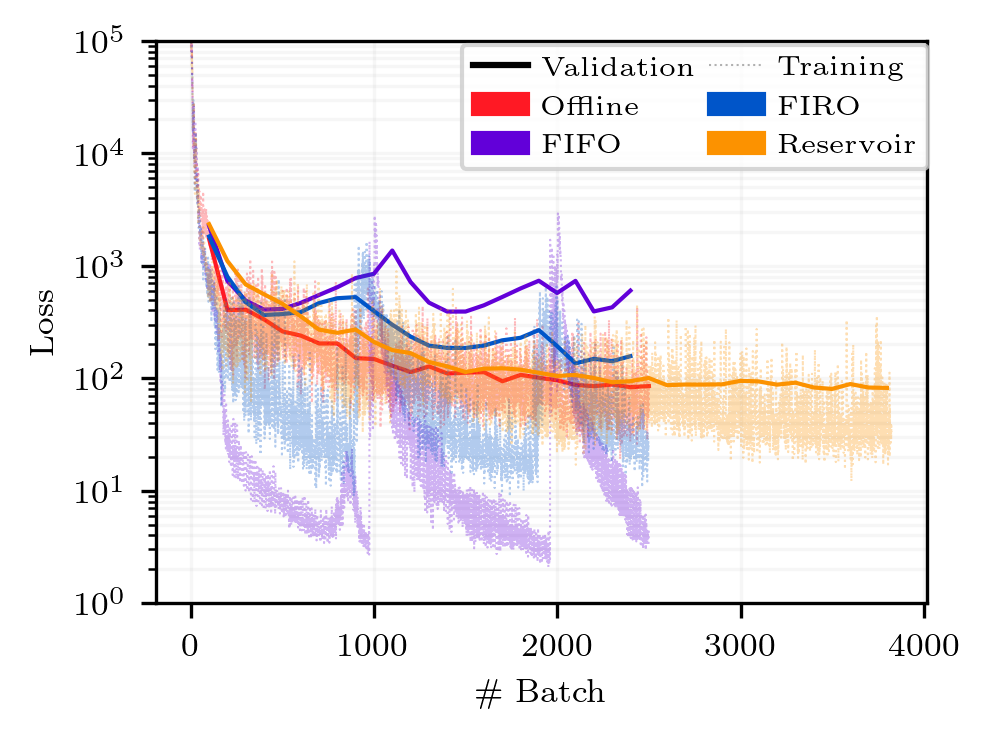}
    \caption{Comparison between training and validation losses for different buffer implementations.}
    \label{fig:losses}
\end{figure}

\autoref{fig:losses} compares the training and validation losses for the different buffers. It also includes scores for offline training
performed over one epoch with data read from files (data are seen only once). In all the different settings, the same unique time steps are seen during training. They only differ by how these time steps are ordered in training batches. For the same amount of unique data, offline training, whose batches are uniformly drawn from the full dataset, does not suffer the biases online methods experience. It is thus expected to provide the best achievable training quality and serve as a reference.

The training data generation replicates the process of the previous experiment (\autoref{sec:exp:throughput}). The same validation dataset is used for all buffers. It consists of 10 simulations generated offline and never seen during training. Validation loss assesses the {\bf generalization capabilities} achieved through training. Because validation occurs on the training thread, it stalls the consumption of batches. Concurrently, incoming time steps are still being processed by the data aggregator thread to fill the buffer. As such, validation is a measure that impacts the experiment it measures. To mitigate this impact, validation is performed every 100 batches. During validation, new entries in the buffer are blocked by acquiring its mutex. Nonetheless, newly produced data sent by the clients still accumulate in the ZMQ buffer. 

During training, the learning rate, initially set to $1\text{E}^{-3}$, is halved every 1000 batches. FIFO presents a low training loss associated with a high validation loss, which indicates overfitting. Both losses experience bursts, occurring when the learning rate is halved, which is symptomatic of unstable training. Although to a lesser extent, FIRO presents the same problems. On the contrary, Reservoir shows a more stable training not subject to overfitting. It achieves a validation loss that is on par with offline training. 
First, the performance of the FIFO simple data streaming confirms that online data generation indeed leads to a strong data bias that affects the training quality. Second, buffering with random data reads as performed by FIRO and Reservoir is effective in mitigating this bias. We assume the higher and thus more diverse population of Reservoir compared to FIRO (as shown in \autoref{fig:throughput}) explains why this former implementation outperforms the latter in terms of generalization capabilities.

Because Reservoir can repeat samples, it can generate more batches than its counterparts, as it appears in \autoref{fig:losses}. Nonetheless, this higher number of training batches does not necessarily translate into longer training time. \autoref{tab:pre-experiments} indicates that training with Reservoir takes less than 6 minutes and is 45 seconds longer than with FIRO, while Offline training takes more than 1 hour to run.

\subsection{Scaling to multiple GPUs}
\label{sec:multigpus}

This experiment evaluates how the different training buffers behave when increasing the number of GPUs using data distributed parallel training. The data generation, training process, and validation remain the same as in the previous experiment (\autoref{sec:exp:training_quality}). To keep the learning rate out of the impacting factors, its update frequency is scaled according to the number of GPUs to match always the same number of training samples. As previously, the learning rate is halved every 10,000 training samples until it reaches a minimum of \(2.5E^{-4}\). Given a batch size of 10, these updates correspond to 1000, 500, and 250 batches for 1, 2, and 4 GPUs respectively.

\begin{table*}[ht!]
\caption{Comparison of the training and throughput performances for different numbers of  GPUs. All experiments rely on 250 simulations producing 100 GB of data or  25,000 unique samples. Fifty nodes are pre-allocated to run  100 simultaneous simulations, each one using  20 cores. The \textit{Min. RMSE} column indicates the validation loss obtained after training.}
\label{tab:pre-experiments}
\vskip 0.15in
\begin{center}
\begin{small}
\begin{sc}
\begin{tabular}{lccccc}
\toprule
\adjustbox{stack=cc}{Buffer \\ ~} & \adjustbox{stack=cc}{Gpu number \\ (n)} & \adjustbox{stack=cc}{Generation \\ (hours)} & \adjustbox{stack=cc}{Total \\ (hours)} & \adjustbox{stack=cc}{Min. MSE $\downarrow$ \\ ~} & \adjustbox{stack=cc}{Mean. Throughput \\ (samples/sec)} \\
\midrule
    Offline & 1 & 0.22 & 1.13 & 83.1 & 13.2 \\
    FIFO & 1 & \textemdash & 0.0805 & 391 & 118 \\
    FIRO & 1 & \textemdash & 0.0832 & 135 & 114 \\
    Reservoir & 1 & \textemdash & 0.0928 & 80.3 & 147.6 \\
    \midrule
    Offline & 2 & 0.22 & 0.353 & 112 & 30.2 \\
    FIFO & 2 & \textemdash & \textbf{0.0793} & 384 & 105 \\
    FIRO & 2 & \textemdash & 0.0835 & 202 & 98.1 \\
    Reservoir & 2 & \textemdash & 0.0972 & 89.3 & 212 \\
    \midrule
    Offline & 4 & 0.22 & 0.201 & 218 & 43.2 \\
    FIFO & 4 & \textemdash & 0.0799 & 445 & 100 \\
    FIRO & 4 & \textemdash & 0.0824 & 197 & 96.4 \\
    Reservoir & 4 & \textemdash & 0.0952 & \textbf{65.0} & \textbf{476} \\
\bottomrule
\end{tabular}
\end{sc}
\end{small}
\end{center}
\vskip -0.1in
\end{table*}

\autoref{tab:pre-experiments} shows in its last column the average throughput during training for the different buffers and different numbers of GPUs. FIFO and FIRO fail to provide higher throughput when the number of GPUs increases. The population size displayed in \autoref{fig:throughput} was already showing FIFO and FIRO were unable to balance the production of data with the higher consumption. Increasing the number of GPUs only worsens this trend, because it increases the consumption and dilutes the generated data between the buffer replicas. Only the Reservoir scales with the number of GPUs at equal data production. It is important to remember however that because its higher throughput is due to data repetitions, this leads to more batches and thus training can eventually be slightly longer. \autoref{tab:pre-experiments} summarizes the results obtained for the different implementations and different numbers of GPUs. 

\begin{figure}[h]
    \includegraphics[width=\columnwidth]{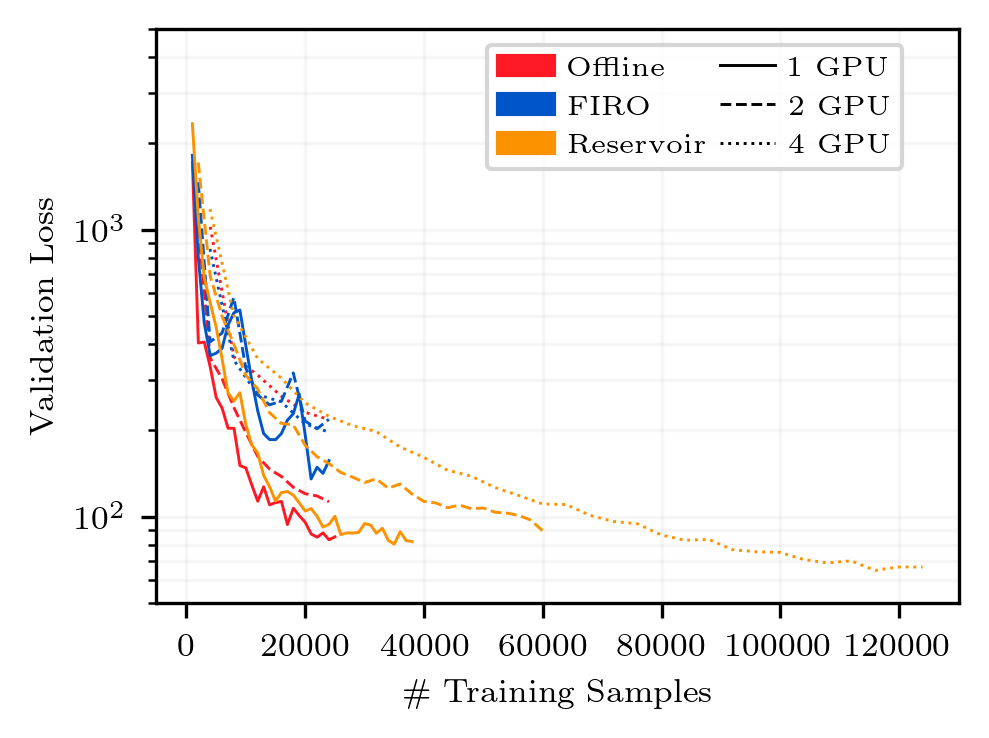}
    \caption{Comparison of the validation loss for different buffer implementations and number of GPUs. Training samples represent the number of simulation time steps, possibly with repetition, seen during training.}
    \label{fig:multigpu_losses}
\end{figure}

\autoref{fig:multigpu_losses} displays the validation loss when training with different numbers of GPUs. Beware that as the number of GPU increases the number of batches decreases, the data generation being always the same. To visually compare the different runs, the x-axis corresponds to the number of simulation time steps seen during training. This number, \(n_s\), is related to the number of batches, \(n_b\), by the relation \(n_s = n_b \times b \times n_{\text{GPU}}\), where \(b\) is the batch size.

As in the previous experiment, the same number of unique samples is used for each training. Therefore, the generalization performance obtained with offline training, as an assumed optimal, still constitutes a benchmark. As the number of GPUs increases the validation loss decreases. This could be explained by the reduced number of optimization steps performed due to a lower number of batches.

The parameters of the buffer, namely the threshold and the maximum capacity, are the same for the different implementations and across the multiple GPUs. So the global buffer storage capacities increase with the number of GPUs, potentially enabling the production of more diverse batches. At an equivalent number of training samples, this greater diversity, however, does not compensate for the deterioration of the validation observed while training on more GPUs. Except for the run with 4 GPUs the validation loss obtained with FIRO is never matching the offline reference. 

For Reservoir, increasing the number of GPUs also increases the global buffer size. But as the data production remains the same, the more GPUs, the less new data each local buffer receives. Besides, batches being larger with more GPUs, more data must be extracted at each training step. Thus, increasing the number of GPUs induces more sample repetition (as seen in \autoref{fig:seen}), which results in more training batches. There are 6 times more batches generated by Reservoir with 4 GPUs compared to 1. These additional batches trigger more optimization steps. The greater number of optimization steps that characterize Reservoir can explain why it beats the one-epoch offline training benchmark, not unlike how training for multiple epochs would improve the offline setting. However, Reservoir repeats samples at a much faster pace than would an offline multi-epoch training (\autoref{tab:pre-experiments}).

Reservoir consistently outperforms all the other training settings for the same number of GPUs. At the end of each training, the validation loss of Reservoir is significantly lower than the one of FIRO, often more than halved. The Reservoir capability to keep its buffer full with a self-adjusted amount of new and already-seen data makes it more amenable to take full benefit of multi-GPU training compared to FIRO. It provides higher throughput and better generalization materialized in a lower validation loss.

\subsection{Online versus multi-epoch Offline}
\label{sec:offline}

\begin{table*}[ht!]
\caption{Comparison with 4 GPUs. The \textit{RESOURCES} column indicates computing resources used for data generation and training. The online experiment trains the network with 20,000 simulations producing 8TB of data or 2,000,000 unique samples. A total of 128 nodes are pre-allocated to run  512 simultaneous simulations, each one using  10 cores.}
\label{tab:scale-experiments}
\vskip 0.15in
\begin{center}
\begin{small}
\begin{sc}
\begin{tabular}{lcccccccc}
\toprule
\adjustbox{stack=cc}{Buffer \\ ~} & \adjustbox{stack=cc}{Generation/Training \\ Resources(\textbf{C}ores \& \textbf{G}pu)~} & \adjustbox{stack=cc}{Generation \\ (hours)} & \adjustbox{stack=cc}{Total \\ (hours)} & \adjustbox{stack=cc}{Dataset\\ (GB)} & \adjustbox{stack=cc}{Unique Samples \\ ($N$)} & \adjustbox{stack=cc}{MSE $\downarrow$ \\ ~} & \adjustbox{stack=cc}{Throughput \\ (samples/sec)} \\
\midrule
    Offline & 2,000C / 40C, 4G & 0.22 & 24.5 & 100 & 25,000 & 25.1 & 38.2\\
    Reservoir & 5,120C / 40C, 4G & \textemdash & 1.97 & 8,000 & 2,000,000  & 13.2 & 476.7 \\
\bottomrule
\end{tabular}
\end{sc}
\end{small}
\end{center}
\vskip -0.1in
\end{table*}

\begin{figure}[h]
    \includegraphics[width=\columnwidth]{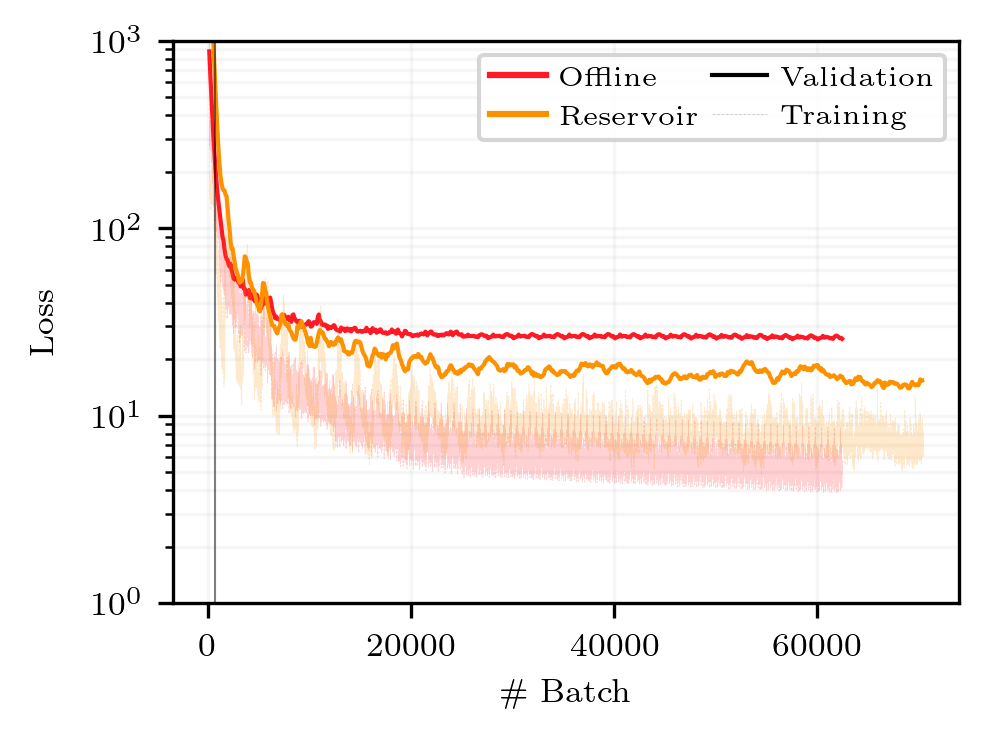}
    \caption{Comparison between offline and online training at equivalent numbers of batches. Offline trains for several epochs on the data generated by 250 simulations. The black vertical line corresponds to the first epoch. Online training is performed on 20,000 simulations managed by the framework.}
    \label{fig:large_scale}
\end{figure}


We have seen that properly managed online training, with Reservoir being the best option, is already competitive with offline training on a single epoch. But the true potential of online training comes when enabling working with datasets so large that they cannot be reasonably handled in an offline fashion, due to storage and I/O costs. Offline needs to restrain to a reasonable dataset presented several times through different epochs, while online can lean towards training on a potentially unlimited dataset. In the following, we compare offline and online in such a context.

Offline training is performed on the same 250 simulations data as used for previous experiments, except that here we perform 100 epochs (\autoref{fig:large_scale}).  The dataset is  450 GB raw with one file per time step, 95.5 GB when compressed using one binary file per simulation. The data generation is also performed in parallel with the framework using 2000 cores, but instead of streaming the data through the API, they are simply written on disk. Here, the framework reveals itself also useful to quickly generate datasets by leveraging the parallelism of its clients. During training, the data are loaded from SSD disk. The loading relies on \emph{mmap} to read only the requested time sep without having to load the entire file in memory. On each of the 4 GPUs synchronized with Pytorch Distributed and individually associated to 10 cores, the Dataloader retrieves batches with 8 parallel workers. Although more advanced techniques exist for fast data loading of massive deep learning datasets \cite{aizmanHighPerformanceLarge2019,zhuPHDFSOptimizingPerformance2020,paulCharacterizingMachineLearning2021,dryden2021clairvoyant}, we believe the current process is fairly representative of common practice in the deep learning community.




Online training is performed with  20,000 simulations managed by the framework. The simulation clients run on  5,120 cores and generate a total of 8TB that are forwarded as soon as produced to the training server running with 4 GPUs.

\autoref{fig:large_scale} compares the training and validation losses for the two settings. The offline training displays clear signs of overfitting with a validation loss that has converged to a higher value than the training loss that keeps decreasing. Overfitting is less pronounced with the Reservoir as both validation and training losses keep decreasing. Overall, Reservoir training significantly improves the validation loss, and so the surrogate generalization capabilities, by $47\%$. 

\autoref{tab:scale-experiments} summarizes the parameters of these two training settings, including throughput and execution times. Using both 4 GPUs, offline only manage to process about $38$ samples/sec, even when using $8$ data loaders per GPU, incurring a training of $24$h. Online Reservoir training enables to process $476$ samples/sec, leading to a combined data generation and training in less than $2$h.

\section{Conclusion}
\label{sec:conclusion}

In this paper, we presented the Melissa framework and introduced the Reservoir algorithm. The former enables online training from an ensemble of simulation runs. The latter is a training buffer that adequately mitigates the bias inherent to online learning while optimizing throughput. The combination of both allows high-quality training of deep surrogates. Experiments revealed that GPUs can indeed process batches at high frequency, way beyond what is achieved with offline training. Using consolidated figures provided by the supercomputer center ($1$ kh/core CPU = $6$\texteuro, $1$ kh/GPU V100 = $360$\texteuro,  $1$TB (SSD storage) = $56$\texteuro), leads to a cost of our large experiment (\autoref{tab:scale-experiments}) with online training at $63.8$\texteuro, only $29\%$ above the cost of offline data generation and training at $49.1$\texteuro.  The cost of offline training would decrease to $41.16$\texteuro\  when repeated (no storage and data generation costs). If offline training would have been performed with the $8$TB dataset of online training, the sole storage cost would account for $480$\texteuro.  A realistic production workflow will likely combine pre-training (with the necessary repetitions to tune hyperparameters) from a static reduced dataset and few online re-training at scale with complementary data so as to reach the best possible generalization capabilities. This will enable to control the trade-off between storage footprint and the computing cost of re-running simulations for each new training.






The present work does not include the use of the surrogate. Thus, the global gain, counting the cost of training and use of the surrogate, versus using only the original solver cannot be evaluated. But higher generalization capabilities mean a surrogate capable of giving higher quality results, likely leading to reduce surrogate runs as well as some of the simulation runs that are often required to bring higher quality data into the process.

Experiments conducted for this paper (including preparatory and test runs)  account for $584062$ core.h and $4770$ gpu.h, leading to $1.1$TCO2e emissions (counting direct energy use, operating costs and hardware construction), or $44\%$ of the one person round-trip flight from Paris to Denver in economy class that presenting this paper will require.

The framework can support adaptive training where the next set of clients to run is defined online according to the current training status. This could increase generalization capabilities while requiring fewer simulations to run. It is only possible in the online context the framework provides. This will be the object of future investigations.

\begin{acks}
This work was performed using HPC/AI resources from GENCI-IDRIS (Grant 2022-[AD010610366R1]),
and received funding from the European High-Performance Computing Joint Undertaking (JU) under grant agreement No 956560. 
\end{acks}

\appendix
\section{Reservoir}
\begin{proof} 
\label{proof:item_life_expectancy}

Let's consider a container of fixed capacity $n$ on which $m$ new items are
sequentially added, with $m \gg n$. New items are inserted at random locations,
overwriting any data already present in the container at these locations.
Considering a new insertion, the probability for an already present item to
remain in the container is the probability to select any location other than the
one where the item lies \emph{i.e} $\frac{n-1}{n}$. Thus, for an item to remain
in the buffer after $k$ insertions the probability is defined by
\autoref{eq:proof1:1}. The factor $\frac{1}{n}$ comes from normalization so to
have $\sum_{k=0}^{+\infty}p(k)=1$.

\begin{equation}
    \label{eq:proof1:1}
    p(k) = \frac{1}{n}(1 - \frac{1}{n})^k
\end{equation}

The expected residency time $\tau$ in the container for any item is then given by
$\sum_{k=0}^{+\infty}k\cdot p(k)$.

\begin{align}
    \tau &= \frac{1}{n} \sum_{k=0}^{+\infty}k(1-\frac{1}{n})^k \label{eq:proof1:2}\\
    &= n - 1 \label{eq:proof1:3}
\end{align}

The step from \autoref{eq:proof1:2} to \autoref{eq:proof1:3} is made by
recognizing in \autoref{eq:proof1:2} the derivative of a converging geometric
series.

To intuitively understand this result, one can consider, as the insertion is
done at a random location, it is as if the eviction of old items is performed
sequentially. Hence, an expected residency time for any item of $n-1$.
\end{proof}

\bibliographystyle{ACM-Reference-Format}
\bibliography{references}

\end{document}